\documentclass[lettersize,journal]{IEEEtran}
\usepackage{amsmath,amsfonts}
\usepackage{algorithmic}
\usepackage{algorithm}
\usepackage{array}
\usepackage[caption=false,font=normalsize,labelfont=sf,textfont=sf]{subfig}
\usepackage{textcomp}
\usepackage{stfloats}
\usepackage{url}
\usepackage{verbatim}
\usepackage{graphicx}
\usepackage{cite}
\usepackage{amssymb}
\usepackage{booktabs}
\usepackage[table,xcdraw]{xcolor}
\usepackage[normalem]{ulem}
\usepackage{multirow}
\usepackage{booktabs}
\usepackage{makecell}
\usepackage{xcolor}
\usepackage[table]{xcolor}
\useunder{\uline}{\ul}{}
\usepackage[hidelinks]{hyperref}

\definecolor{bestred}{RGB}{255,225,225}
\definecolor{basegray}{RGB}{230,234,250}
\definecolor{gainred}{RGB}{190,80,80}
\definecolor{baseindigo}{RGB}{230,234,250}
\newcommand{\best}[1]{%
  \begingroup
  \setlength{\fboxsep}{1.2pt}%
  \colorbox{bestred}{\textbf{#1}}%
  \endgroup
}

\begin{document}

\title{Adaptive Inference-Time Scaling via Early-Step \\ Latent Verification for Image Editing}

\author{Yue~Yu,
        Yang~Jiao,
        Jiayu~Wang,
        Qi~Dai,
        Jingjing~Chen,~\IEEEmembership{Senior Member,~IEEE}
\thanks{Yue Yu, Yang Jiao, and Jiayu Wang are with the College of Computer
Science and Artificial Intelligence, Fudan University, Shanghai 200433, China (e-mail: {yuy24, jiayuwang25, yjiao23}@m.fudan.edu.cn).}
\thanks{Qi Dai is with Microsoft Research Asia, Beijing 100080, China (e-mail: {qid}@microsoft.com).}
\thanks{Jingjing Chen is with Institute of Trustworthy
Embodied AI, Fudan University, Shanghai 200433, China (e-mail: {chenjingjing}@fudan.edu.cn).}
\thanks{This work has been submitted to the IEEE for possible publication. Copyright may be transferred without notice, after which this version may no longer be accessible.}
}

\markboth{Preprint.}%
{Shell \MakeLowercase{\textit{et al.}}: A Sample Article Using IEEEtran.cls for IEEE Journals}


\maketitle

\begin{abstract}
Instruction-based image editing has made notable progress with recent advances in generative models. However, the quality of the edited result is still influenced by the randomly sampled initial noise, particularly in complex editing scenarios. An unsuitable initial noise may lead to unsatisfactory editing results. Recent inference-time scaling methods address this issue by sampling multiple initial noises and selecting better candidates. Nevertheless, most of them follow a decode-then-verify scheme which introduces an efficiency-accuracy trade-off. When decoding is performed after limited inference steps, the decoded images often remain too noisy for reliable assessment, whereas sufficiently denoised images require much higher computational cost. To address this issue, we propose VeriLatent, a plug-and-play adaptive inference-time scaling framework with early-step latent verification for image editing. Specifically, we propose a novel verifier that scores each initial noise through a latent-space editing activation map at an early stage. It identifies promising candidates by assessing whether they can induce an effective edit in the correct region. This enables efficient early pruning without decoding latents into images. Building on this, we further develop an adaptive search strategy for inference-time scaling. It allocates inference budgets according to editing difficulty, thereby reducing the number of function evaluations (NFE). Extensive experiments on multiple benchmarks and different base models demonstrate that VeriLatent consistently improves both editing performance and inference-time scaling efficiency. Code is released at \textcolor{magenta}{\textit{\url{https://github.com/Yue-105/VeriLatent}}}.
\end{abstract}

\begin{IEEEkeywords}
Image editing, instruction-based image editing, inference-time scaling, image CoT.
\end{IEEEkeywords}

\section{Introduction}
\IEEEPARstart{I}{nstruction-based} image editing is an important task in visual content generation and has attracted increasing attention due to its broad practical applications. It aims to modify a given image according to textual instructions. This task covers a wide range of editing scenarios, such as changes to objects, attributes, backgrounds, and styles. However, handling complex editing cases remains a challenging problem, as existing methods cannot consistently produce satisfactory editing results.


Before recent progress in generative models, earlier image editing methods were mainly based on GANs \cite{goodfellow2014generative, mirza2014conditional} and for specific editing settings \cite{isola2017image, zhu2017unpaired, he2018chipgan, he2019attgan}. With the rapid development of image generative models \cite{ho2020denoising, rombach2022high,podell2023sdxl, zhou2026semantic}, researchers have successfully adapted various diffusion-based models for instruction-based image editing. \cite{kawar2023imagic, mokady2023null} perform test-time finetuning to enable customized editing of a given image. Other methods adopt a training-free paradigm and achieve editing through control mechanisms such as inversion, attention modification, and mask guidance. Furthermore, training-based works, such as InstructPix2Pix \cite{brooks2023instructpix2pix} and Emu Edit \cite{sheynin2024emu}, leverage large-scale instruction-based editing data to achieve stronger performance. Recent advanced editing models, including FLUX.1 Kontext, Step1X-Edit \cite{liu2025step1x}, and Qwen-Image-Edit \cite{wu2025qwen}, also follow this training-based paradigm and demonstrate superior performance across diverse editing tasks.

\begin{figure}[t]
    \centering
    \includegraphics[width=8.8cm]{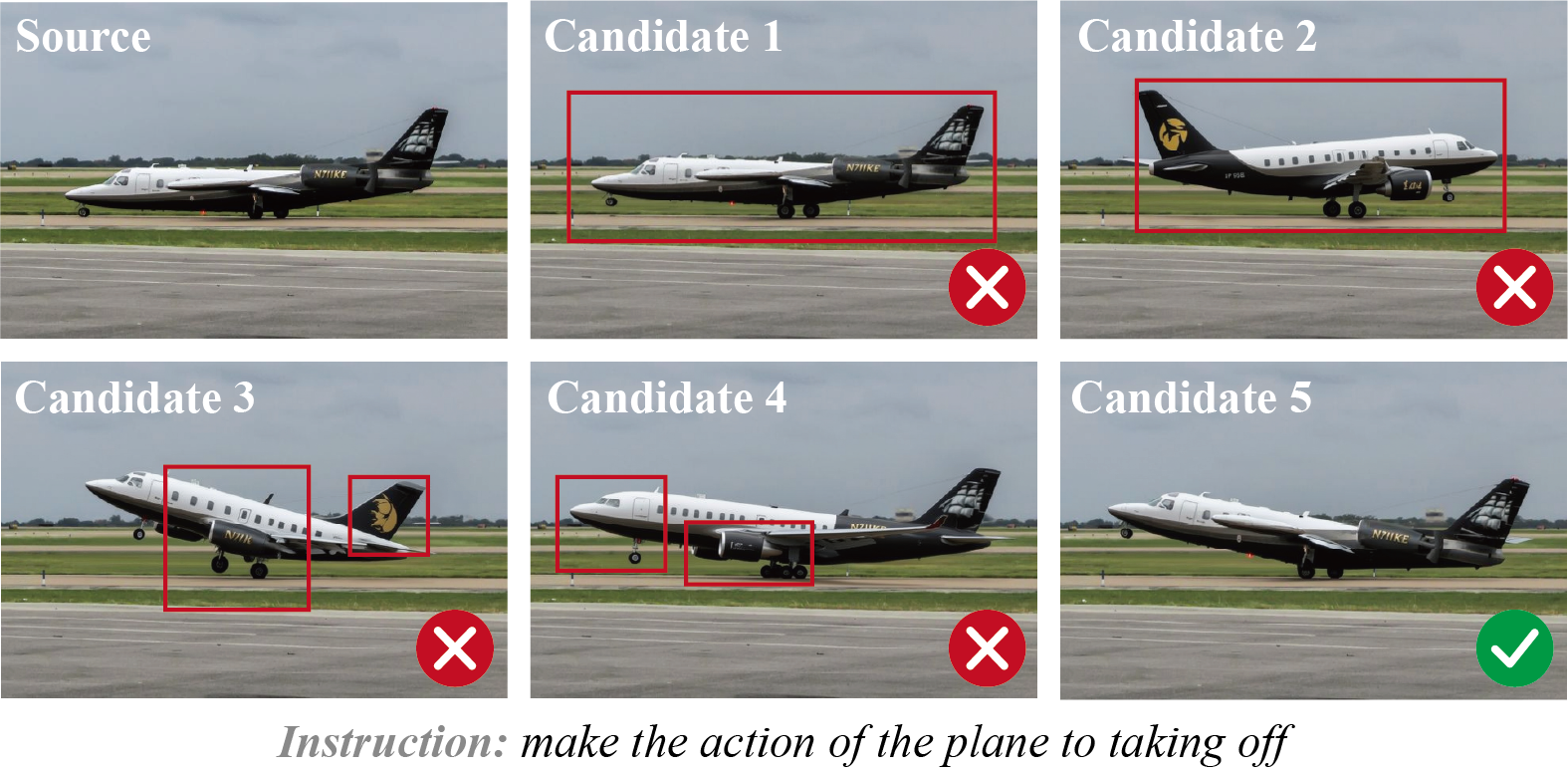}
    \caption{Effect of initial noise on instruction-based image editing. Given the same source image and instruction, FLUX.1 Kontext produces different editing results when using different initial noises. Only Candidate 5 produces a satisfactory result, while the other candidates suffer from ineffective edits or unwanted changes.}
    \label{first}
    \vspace{-0.5cm}
\end{figure}

Despite the recent progress of these methods, instruction-based image editing remains challenging in complex scenarios, as existing methods still cannot consistently produce satisfactory results. A key reason for this instability lies in the stochastic nature of diffusion-based and flow-based generative models, whose generation process starts from randomly initialized noise. As shown in Fig.~\ref{first}, given the same input, the same model can produce different editing results when using different initial noises. An unsuitable initial noise may degrade the editing quality or even lead to failure, especially for complex cases. Therefore, obtaining better initial noise can serve as an effective way to enhance the performance of image editing models. To this end, \cite{ma2025inference} first introduced inference-time scaling into image generation. It formulates diffusion sampling process as a search problem for identifying better initial noises. Subsequent work \cite{zhang2026enabling, guo2025can, singhal2025general, qu2026scale} further extends this line of research to image editing by improving the search strategy and verifier design. Although these methods improve editing performance, their verification scheme remains limited. Most existing methods adopt a \textit{decode-then-verify} scheme. For each candidate initial noise, they first perform a certain number of denoising steps and then decode the insufficiently denoised latent into a preview image for verification. This scheme introduces an efficiency–accuracy trade-off. Since image-level verifiers are typically designed for clean images, such a preview image may remain noisy and incomplete, leading to inaccurate evaluation. Conversely, generating a cleaner preview image requires more denoising steps and thus requires higher computational cost.

To address this issue, we propose an \textbf{early-step latent verifier} that assesses candidate initial noises without decoding latents into pixel space. Specifically, this verifier estimates editing activation at early denoising stage, which reflects how strongly each latent token is likely to be edited. The estimated activation allows the verifier to judge whether the editing tendency is aligned with the instruction. By verifying directly in latent space instead of decoding latents into preview images, the verifier can identify promising initial noises using only a few denoising steps, thereby reducing computational cost. Built upon the early-step latent verifier, we further propose \textbf{VeriLatent}, a plug-and-play inference-time scaling framework for instruction-based image editing. The framework operates in two stages. In the first stage, VeriLatent explores the joint space of initial noises and editing instructions, which allows it to handle ambiguous or complex editing commands. Concretely, it uses the proposed verifier to prune candidates accurately and efficiently at an early stage. At the same time, the exploration budget is allocated adaptively according to the editing difficulty of each instance, thereby further improving computational efficiency. In the second stage, the candidates retained after pruning are further denoised, and the final output is selected from them based on instruction alignment and visual quality.

To summarize, our contributions are threefold:

\begin{itemize}
    \item We propose a novel early-step latent verifier that enables accurate and efficient assessment of candidate initial noises at the early stage of the denoising process. It operates directly in latent space without decoding to pixel space.
    \item We introduce VeriLatent, a plug-and-play inference-time scaling framework for instruction-based image editing, which adaptively explores the joint space of initial noises and editing instructions and efficiently selects high-quality outputs.
    \item Extensive experiments across multiple benchmarks and base models demonstrate that our approach consistently improves instruction-following performance while reducing computational cost. Compared with Best-of-N (BoN), it reduces NFE by approximately 5.5–6$\times$.
\end{itemize}

\section{Related Work}

\subsection{Instruction-based Image Editing}
Instruction-based image editing aims to modify an image according to a given text prompt. With the rapid development of diffusion models \cite{song2019generative, ho2020denoising, rombach2022high, podell2023sdxl}, image generation models have achieved substantially stronger generative capability across diverse generation scenarios \cite{zhang2023adding, xu2024sgdm, zhang2024mmginpainting, gu2026compositional}. Benefiting from this progress, a variety of approaches have been explored for instruction-based image editing. \cite{kawar2023imagic, mokady2023null} adopt test-time finetuning, where the model is optimized on the input image during inference for image-specific editing. In addition, \cite{miyake2025negative, huberman2024edit, wallace2023edict} are inversion-based methods. They first invert the input image to a noisy latent and then use this latent as the starting point to generate the edited results. Another common category of methods is attention modification \cite{hertz2022prompt, tumanyan2023plug, cao2023masactrl, xu2023inversion}. These methods control the generation process by manipulating attention maps and feature maps.

Beyond these methods, training-based methods have become a more effective paradigm for instruction-based image editing, as they directly learn editing behaviors from large-scale instruction-based data. Early works, such as InstructPix2Pix \cite{brooks2023instructpix2pix}, Emu Edit \cite{sheynin2024emu}, and MagicBrush \cite{zhang2023magicbrush}, are trained on curated datasets, enabling the model to acquire instruction-following capabilities. With the advancement of flow-based generative architectures \cite{liu2022flow, esser2024scaling}, training-based methods have been further enhanced for instruction-based image editing. Recent advanced models, including FLUX.1 Kontext, Step1X-Edit \cite{liu2025step1x}, and Qwen-Image-Edit \cite{wu2025qwen}, have demonstrated strong performance across diverse editing tasks. Nevertheless, even these advanced methods remain limited in complex editing cases. For difficult instructions, e.g., large pose changes, motion changes, or multi-object editing, the generated results are strongly influenced by the randomly sampled initial noise. Only a small subset of noises can lead to effective edited results. As a result, these models cannot consistently produce satisfactory edits. Our proposed method aims to address this problem by identifying suitable initial noises via inference-time scaling.

\subsection{Inference-time Scaling}
The concept of inference-time scaling originates from inference strategies in large language models \cite{snell2024scaling, brown2024large, gandhi2024stream, yao2023tree}. This approach improves the quality of generated responses by scaling up the computation during inference through complex search processes. In addition, \cite{ahn2024noise,qi2024not} have shown that the quality of noise used in diffusion models is different, which affects the quality of the generated images.
Building on this insight, \cite{ma2025inference} introduces this paradigm to diffusion models. By employing appropriate verifiers and noise search algorithms, it is possible to obtain higher-quality noise, thereby improving the quality of generated images. \cite{zhang2025let} adapts Chain-of-Thought (CoT) reasoning to autoregressive image generation on Show-o \cite{xie2025show}. They propose PARM, which employs two verifiers for candidate pruning. A clarity verifier first determines whether an intermediate state is sufficiently clear for assessment, and a potential verifier then judges whether the corresponding path is worth retaining. 

Recent works have also explored inference-time scaling for instruction-based image editing. ELECT \cite{kim2025early} improves instruction-based image editing by selecting reliable seeds based on background consistency at intermediate stage of denoising (e.g., about 40 out of 100 denoising steps). ICEdit \cite{zhang2026enabling} first runs a fast inference with a small number of steps for multiple initial noises and decodes the preliminary outputs. It then adopted a MLLM verifier to select the optimal initial noise for full-step inference. ADE-CoT \cite{qu2026scale} further improves candidate selection with a difficulty-aware adaptive strategy. Overall, most existing methods adopt a decode-then-verify scheme. This scheme introduces an efficiency-accuracy trade-off. With a reduced number of inference steps, the decoded image often remains noisy. This makes verification unreliable, since most verifiers are designed for clean images. In contrast, more inference steps improve verification accuracy but lead to a higher computational cost. To address this issue, we propose a novel verifier that directly evaluates early-step latents without decoding.


\section{Method}

\subsection{Preliminaries}
In instruction-based image editing, recent editing models are commonly built on rectified flow \cite{lipman2022flow, liu2022flow}. Given an input image $I_\mathrm{src}$ and an editing instruction $P$, the model generates an edited image by transporting an initial noise sample $x_1 \sim p_1$ toward the target edited sample $x_0 \sim p_0$ through a fixed number of denoising steps, where $p_1$ denotes the Gaussian noise distribution and $p_0(\cdot \mid I_{\mathrm{src}}, P)$ denotes the conditional distribution of edited latents. Rectified flow formulates this transport process as a continuous trajectory governed by an ordinary differential equation (ODE):
\begin{equation}
    \mathrm d x_t = v_{\theta}(x_t, t, I_\mathrm{src}, P)\mathrm dt,
\end{equation}
where $v_{\theta}$ is the predicted velocity and $x_t$ denotes the intermediate latent at time $t$. With this ODE, an initial noise $x_1$ can be transformed into the edited latent $x_0$, which can then be decoded into the edited image.

\subsection{Early-step Verifier in Latent Space}
For instruction-based image editing, there are three common failure modes: (1) no effective edit is triggered, leaving the output almost unchanged; (2) the edit occurs in incorrect regions rather than the target regions; and (3) the generated image has degraded visual quality, such as visible artifacts. We observe that the first two failure modes can often be detected in the early denoising stage in the latent space without decoding. Motivated by this and to address the efficiency-accuracy trade-off issue, we propose an early-step latent verifier to assess the quality of initial noises.

\subsubsection{Editing Activation and Region Detection}
The proposed verifier assesses each initial noise from two aspects: editing activation and edit-region accuracy. Editing activation measures whether the initial noise can induce an effective edit. To determine this, we first conduct denoising from initial noise to an early stage $t_e$, where $(1 - t_e)$ is much smaller than $1$. After this stage, we use the flow-matching formulation to obtain an estimated previewed final latent $\hat{x}_{0}$ through a one-step prediction:
\begin{equation}
    \hat{x}_{0} \;=\; x_{t_e} - t_e \cdot v_\theta(x_{t_e}, t_e, I_\mathrm{src}, P),
\end{equation}
Subsequently, we define the token-wise editing activation map of the latent representation as follows:
\begin{equation}
    \mathrm{Activation}_i = 1 - 
    \frac{\hat{x}_{0,i}^{\top} x_{\mathrm{src},i}}{\|\hat{x}_{0,i}\| \, \|x_{\mathrm{src},i}\|},
    \quad i = 1,\dots,L,
\end{equation}
where $x_\mathrm{src}$ is encoded latent of the input image $I_\mathrm{src}$, $i$ indexes the latent tokens, and $L$ denotes the number of latent tokens. The editing activation value reflects the strength of change in each latent token from the input image. For a qualified initial noise, tokens in the target region should have substantially higher activation values than tokens in other regions. If the activation values are low and diffusely distributed, the output image is likely to remain almost unchanged. This is a common failure case for hard editing cases.

\begin{figure}[t]
    \centering
    \includegraphics[width=8.8cm]{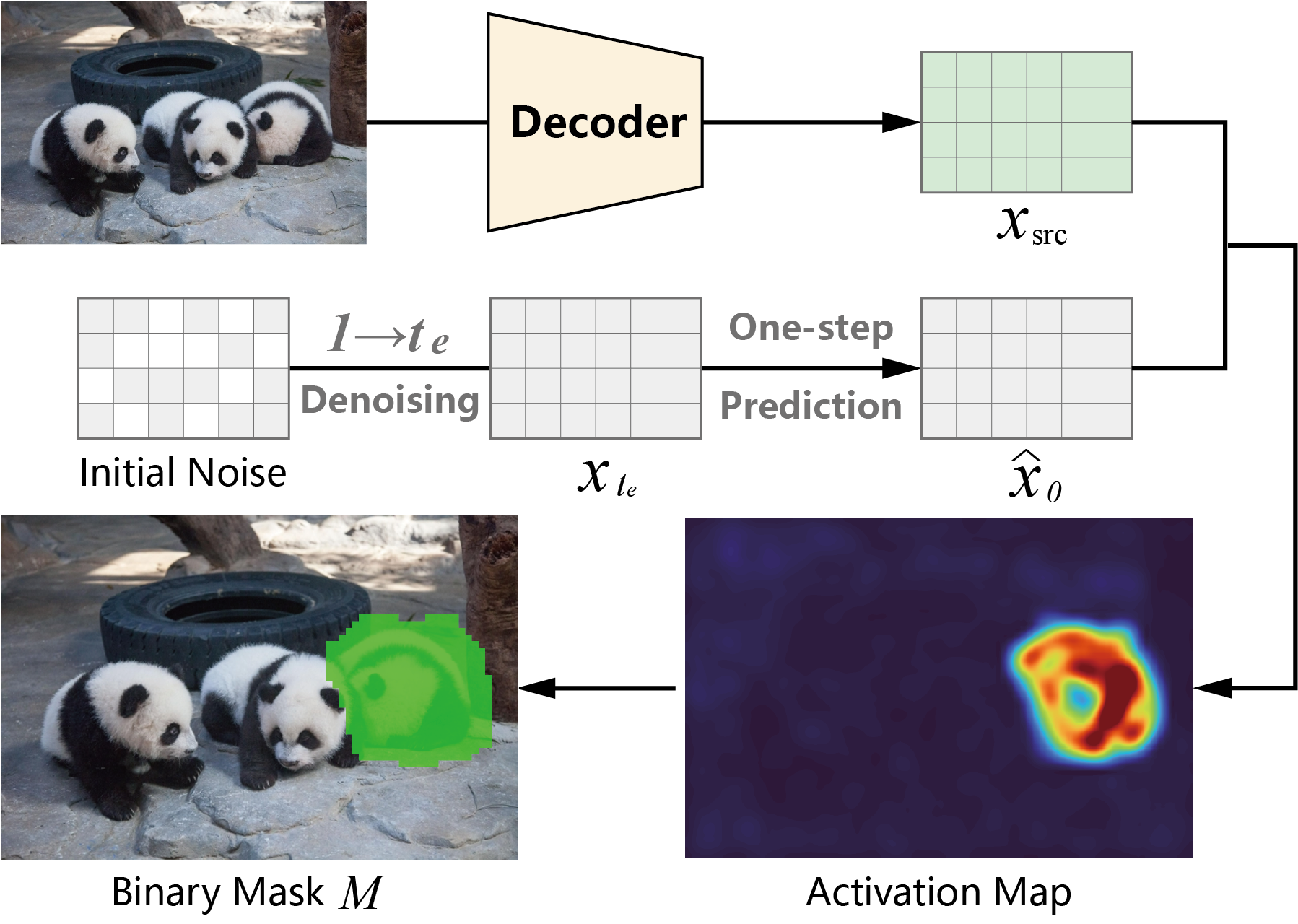}
    \caption{The process of obtaining the token-wise editing activation map and the corresponding binary edit-region mask. The mask is visualized as an overlay on the input image.}
    \label{mask}
    \vspace{-0.5cm}
\end{figure}

Based on the token-wise editing activation map, we can identify the editing region induced by the initial noise. Specifically, tokens with activation values above a threshold are selected as edit-region tokens, while the others are ignored. We further apply morphological operations to connect neighboring edit-region tokens and remove isolated noisy tokens. This produces a cleaner and more coherent binary edit-region mask $M$. Fig.~\ref{mask} demonstrates an example of this process.

\subsubsection{Initial Noise Quality Judge}
The binary edit-region mask provides a direct cue for assessing the initial noise. If the corresponding mask is well aligned with the editing instruction, it suggests that the initial noise can trigger effective editing in the correct region. In contrast, if the corresponding mask is absent, misses the target region, or covers irrelevant regions, the initial noise is less desirable. Therefore, we use the obtained mask to make a preliminary assessment of the initial noise quality. Since this assessment requires understanding both the image content and the editing instruction, we employ a MLLM as a judge to comprehensively evaluate the mask validity from multiple perspectives.

Specifically, for each candidate initial noise and their corresponding mask, we render a side-by-side composite image: the input image on the left, and the same image with the mask overlaid as a translucent green region on the right. The MLLM judge receives the composite, the editing instruction, and the scoring rubric. The rubric covers three aspects: (a) coverage of the target regions, (b) avoidance of irrelevant regions, and (c) appropriateness of the coverage scale (e.g., for global edits such as style transfer, the mask should cover most or all of the image). In particular, we introduce an {\bf empty-mask gating} mechanism. If the mask is empty, no token is activated. In this case, the MLLM may still assign a high score due to its own hallucination. To suppress this issue, we multiply the score by $\mathbb{I}\left[|| M || > 0\right]$, which sets the score to zero for empty masks. This eliminates a clear reward-hacking failure mode observed in our experiments.

Overall, the proposed verifier enables an effective and efficient assessment of initial noise quality in the latent space. Compared with judging based on decoded images, mask-based evaluation is more stable because it is less affected by residual noise caused by insufficient denoising. Therefore, the verifier can assess initial noise quality reliably with only a few denoising steps.

\begin{figure*}[t]
    \centering
    \includegraphics[width=18cm]{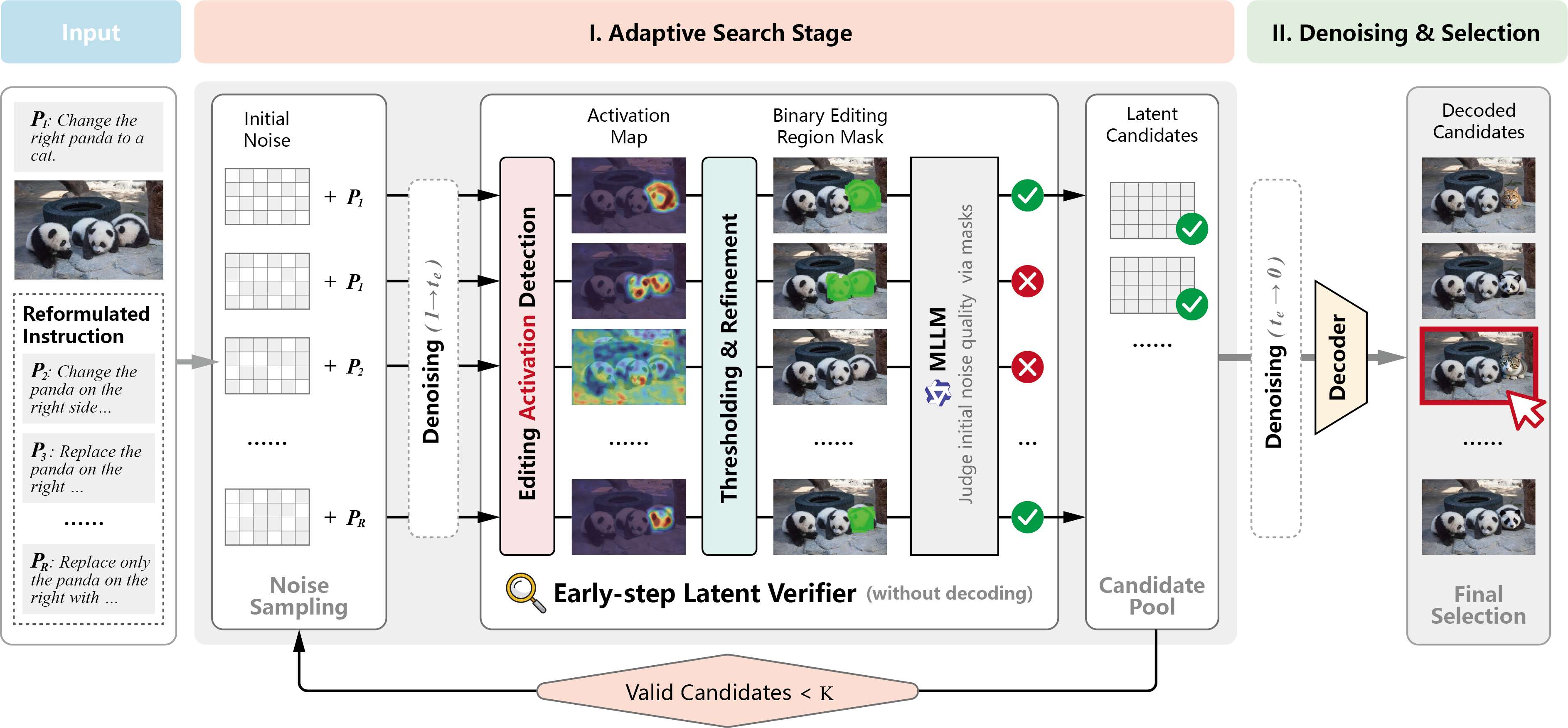}
    \caption{Overview of the framework. Given an input image and an editing instruction, VeriLatent operates in two stages: (1) \textbf{Adaptive Search Stage}: VeriLatent reformulates the instruction, samples candidate initial noises, and prunes unpromising candidates in latent space with an early-step latent verifier; (2) \textbf{Denoising and Final Selection}: it fully denoises the retained candidates and utilizes an additional MLLM to select the most favorable output.}
    \label{framework}
\end{figure*}

\subsection{Inference-time Scaling Framework}
By integrating the proposed early-step verifier,  we propose \textbf{VeriLatent}, a plug-and-play inference-time scaling framework for instruction-based image editing. The framework expands the inference-time computation to explore the joint space of \emph{initial noises} and \emph{editing instructions}, thereby identifying more promising candidates and improving the final editing quality. The proposed framework consists of two stages: an adaptive search stage and a subsequent denoising stage. The overall framework is illustrated in Fig.~\ref{framework}.


\subsubsection{Adaptive Search Stage}
The goal of this stage is to search and assess the quality of candidate initial noises early in the inference process, thereby deciding which candidates should be fully denoised. The search is conducted over both the editing instruction and the initial noise, since the editing result is sensitive to the instruction wording: ambiguous references or overly complex phrasing may lead to ineffective or incorrect edits, even under the same initial noise. To this end, we first adopt a MLLM to rewrite the original instruction conditioned on the input image $I_\mathrm{src}$. The rewriting process reformulates the instruction into a direct and plain editing command that is easier for the editing model to follow, while preserving the original editing semantics. The original instruction is also retained as a fallback to reduce the risk of semantic drift. This yields a candidate instruction set $\{P_1, \dots, P_R\}$ of size $R$, where $P_1$ is the original instruction.

Together with these reformulated instructions, we search over the initial-noise space in an adaptive, multi-round manner. Editing difficulty varies markedly across instances. Some inputs yield a correctly localized edit with almost any initial noise, while others require many attempts. Therefore, we set a maximum budget of $N$ initial noises and spend it adaptively across rounds. In each round, we sample a batch of $B$ initial noises and allocate them evenly across the $R$ instructions. For every candidate, we run $\tau_{e}$ denoising steps and apply the proposed early-step latent verifier to obtain a score. After each round, candidates are ranked by their verification scores, and candidates above the score threshold $s_{\text{th}}$ are selected. The search terminates once the number of such candidates reaches a predefined number $K$. If the budget is exhausted before $K$ candidates meet the criterion, we retain the $K$ highest-scoring candidates found so far. The retained candidates are then passed to the subsequent denoising stage. This strategy avoids unnecessary computation on easy instances and allocates additional attempts to hard instances. The full procedure is summarized in Algorithm \ref{alg:search}.

\begin{algorithm}[t]
    \caption{The Proposed VeriLatent Framework}
    \label{alg:search}
    \begin{algorithmic}[1]
    \REQUIRE input image $I_\mathrm{src}$, instruction $P_1$; per-round batch size $B$, early denoising step number $\tau_e$, retained pool size $K$, threshold $s_\mathrm{th}$, budget $N$
    \STATE $\{P_1,\dots,P_R\} \leftarrow \textsc{Reformulation}(I_\mathrm{src}, P_1)$
    \STATE $\mathcal{C} \leftarrow \emptyset$, $n \leftarrow 0$
    \WHILE{$n < N$}
        \STATE Sample a batch of $B$ initial noises $\{\epsilon_i\}_{i=1}^{B}$
        \STATE Construct candidate pairs $\mathcal{B}=\{\langle \epsilon_i, P_r \rangle\}$ by assigning noises evenly to the $R$ instructions
        \STATE Run $\tau_e$ denoising steps for each pair
        \STATE Score each candidate using the \textbf{early-step latent verifier}
        \STATE $\mathcal{C} \leftarrow \operatorname{TopK}\left(\mathcal{C} \cup \mathcal{B}, K\right)$
        \STATE $n \leftarrow n + B$
        \IF{the lowest score in $\mathcal{C} \ge s_\mathrm{th}$}
            \STATE \textbf{break}
        \ENDIF
    \ENDWHILE
    \STATE Finish denoising for all candidates in $\mathcal{C}$ and decode them
    \STATE \textbf{return} the best in $\mathcal{C}$ selected by the MLLM judge
    \end{algorithmic}
\end{algorithm}

\subsubsection{Denoising Stage and Final Selection}
For the top-$K$ candidates retained by the adaptive search stage, we perform the remaining denoising steps and decode the resulting latents into pixel space. We keep multiple candidates instead of selecting a single one at the early stage. The early-step verifier assesses candidates in latent space, focusing on editing activation and edit-region accuracy rather than the visual quality or fidelity. Thus, we utilize an additional MLLM-based verifier for final selection, which evaluates the retained candidates and selects the most favorable one as the final output. In this way, the two verifications are complementary: the proposed early-step verifier performs efficient latent-space pruning, while the final selection assesses the final instruction alignment and visual quality in pixel space.

\section{Experiments}

\subsection{Experimental settings}
{\bf Implementation Details}. To evaluate the proposed VeriLatent, we apply it to two advanced base editing models, FLUX.1 Kontext-dev \cite{batifol2025flux} and Step1X-Edit (v1.0) \cite{liu2025step1x}, both with 28 inference steps. For the early-step verifier, we run $\tau_e=6$ denoising steps and compute token-wise editing activation; the activation threshold is set to $0.93$ for FLUX.1 Kontext and $0.80$ for Step1X-Edit. For the overall quantitative evaluations reported in Table~\ref{tab:main_comparison}, we set the maximum budget to $N=32$, the per-round batch size to $B=8$, the number of instructions to $R=4$, the number of retained candidates to $K{=}4$, and the score threshold to $s_{\text{th}}{=}12$. We use GPT-5.4 for instruction augmentation, a newly introduced component that operates before generation. For both the early-step latent verification for pruning and the final selection, we use Qwen-VL-MAX \cite{bai2025qwen25vl} following ADE-CoT \cite{qu2026scale} to ensure a fair comparison. All experiments are conducted on a single NVIDIA A100 GPU (80GB).

{\bf Benchmarks and Metrics}. Following the setup of ADE-CoT \cite{qu2026scale}, we conduct experiments on three representative benchmarks and report the corresponding evaluation metrics. GEdit-Bench-EN \cite{liu2025step1x} is a real-world image editing benchmark built from user editing requests collected from the Internet, covering 11 categories of editing tasks. GPT-4.1 \cite{achiam2023gpt} with VIE-Score \cite{ku2024viescore} is adopted to evaluate semantic consistency (G\_SC), perceptual quality (G\_PQ), and overall score (G\_O). AnyEdit-Test \cite{yu2025anyedit} includes five primary editing categories. We report results on the categories suitable for instruction-based image editing, including local, global, and implicit edits. As for evaluation metrics, $\mathrm{CLIP_{out}}$ \cite{radford2021learning} evaluates the semantic alignment between the generated result and the target output caption. $\mathrm{CLIP_{im}}$ and DINO \cite{oquab2023dinov2} measure the visual similarity between the generated result and the given ground-truth edited image. Reason-Edit \cite{huang2024smartedit} focuses on complex understanding and reasoning scenarios. For this benchmark, we report PSNR and LPIPS for visual consistency in unchanged the background regions, and CLIP score for semantic consistency in the foreground regions that should be edited. Furthermore, we measure computational cost via the Number of Function Evaluations (NFE), which quantifies the cumulative compute count during inference. Lower NFE indicates higher inference efficiency.

\begin{table*}[t]
    \centering
    \fontsize{5.2pt}{7.1pt}\selectfont
    \caption{Comparison with inference-time scaling methods on GEdit-Bench-EN, AnyEdit-Test, and Reason-Edit. Base models are highlighted in light blue, and the \best{best result} under each base model is highlighted in bold with a light red background. The table shows that VeriLatent consistently improves performance across benchmarks and metrics while maintaining a lower NFE compared to other scaling methods.}
    \label{tab:main_comparison}
    \setlength{\tabcolsep}{2.8pt}
    \renewcommand{\arraystretch}{1.18}
    \resizebox{\textwidth}{!}{
    \begin{tabular}{l | c | cccc | cccc | cccc}
    \toprule
    \multicolumn{1}{c|}{\multirow{2}{*}{\textbf{Model}}}
    & \multirow{2}{*}{$N$}
    & \multicolumn{4}{c|}{\textbf{GEdit-Bench-EN (Full)}~\cite{liu2025step1x}}
    & \multicolumn{4}{c|}{\textbf{AnyEdit-Test}~\cite{yu2025anyedit}}
    & \multicolumn{4}{c}{\textbf{Reason-Edit}~\cite{huang2024smartedit}} \\
    \cmidrule(lr){3-6} \cmidrule(lr){7-10} \cmidrule(lr){11-14}
    &
    & G\_SC
    & G\_PQ
    & G\_O
    & NFE$\downarrow$
    & CLIP$_{\text{im}}$
    & CLIP$_{\text{out}}$
    & DINO
    & NFE$\downarrow$
    & PSNR
    & LPIPS$\downarrow$
    & CLIP
    & NFE$\downarrow$ \\
    \midrule
    
    \rowcolor{basegray}
    \textbf{FLUX.1 Kontext~\cite{batifol2025flux}}
    & 1  & 6.546 & 7.615 & 6.104 & --  
         & 0.874 & 0.302 & 0.774 & --  
         & 25.135 & 0.066 & 21.361 & --  \\

    w/ BoN~\cite{ma2025inference} 
    & 32 & 7.132 & \best{7.721} & 6.641 & 896 
         & 0.882 & 0.307 & \best{0.784} & 896 
         & 25.657 & 0.054 & 21.635 & 896 \\
    
    w/ PRM~\cite{zhang2025let} 
    & 32 & 7.018 & 7.713 & 6.517 & 560 
         & 0.880 & 0.306 & 0.782 & 597 
         & 25.633 & 0.058 & 21.603 & 693 \\
    
    w/ PARM~\cite{zhang2025let} 
    & 32 & 7.087 & 7.716 & 6.563 & 768 
         & 0.881 & 0.307 & 0.783 & 695 
         & 25.498 & 0.056 & 21.656 & 826 \\
    
    w/ TTS-EF~\cite{zhang2026enabling} 
    & 32 & 6.866 & 7.657 & 6.376 & 348 
         & 0.878 & 0.305 & 0.779 & 348 
         & 25.509 & 0.057 & 21.639 & 348 \\
    
    w/ ADE-CoT~\cite{qu2026scale} 
    & 32 & 7.225 & 7.719 & 6.695 & 418 
         & \best{0.883} & 0.308 & \best{0.784} & 367 
         & 25.755 & 0.053 & 21.663 & 431 \\
    
    \textbf{w/ VeriLatent (Ours)}
    & 32 & \best{7.637} & 7.663 & \best{7.082} & \best{162} 
         & \best{0.883} & \best{0.311} & 0.774 & \best{150} 
         & \best{27.046} & \best{0.044} & \best{22.619} & \best{154} \\
    
    \midrule
    
    \rowcolor{basegray}
    \textbf{Step1X-Edit~\cite{liu2025step1x}}
    & 1  & 6.942 & 7.184 & 6.420 & --  
         & 0.865 & 0.302 & 0.742 & --  
         & 21.443 & 0.106 & 22.463 & --  \\
    
    w/ BoN~\cite{ma2025inference} 
    & 32 & 7.732 & \best{7.485} & 7.157 & 896 
         & 0.877 & 0.308 & 0.765 & 896 
         & 23.301 & 0.087 & 22.750 & 896 \\
    
    w/ PRM~\cite{zhang2025let} 
    & 32 & 7.647 & 7.405 & 7.031 & 648 
         & 0.874 & 0.307 & 0.762 & 565 
         & 23.118 & 0.090 & 22.791 & 664 \\
    
    w/ PARM~\cite{zhang2025let} 
    & 32 & 7.692 & 7.446 & 7.072 & 689 
         & 0.876 & 0.307 & 0.764 & 693 
         & 23.299 & 0.088 & 22.803 & 710 \\
    
    w/ TTS-EF~\cite{zhang2026enabling} 
    & 32 & 7.296 & 7.301 & 6.777 & 348 
         & 0.873 & 0.306 & 0.758 & 348 
         & 22.273 & 0.095 & 22.502 & 348 \\
    
    w/ ADE-CoT~\cite{qu2026scale} 
    & 32 & 7.821 & 7.465 & 7.196 & 434 
         & 0.878 & 0.309 & \best{0.766} & 434 
         & 23.405 & 0.086 & 22.834 & 446 \\
    
    \textbf{w/ VeriLatent (Ours)}
    & 32 & \best{7.911} & 7.469 & \best{7.277} & \best{156} 
         & \best{0.881} & \best{0.312} & 0.764 & \best{159} 
         & \best{24.841} & \best{0.073} & \best{23.226} & \best{150} \\
    
    \bottomrule
    \end{tabular}
    }
\end{table*}

\begin{figure*}[t]
    \centering
    \includegraphics[width=18cm]{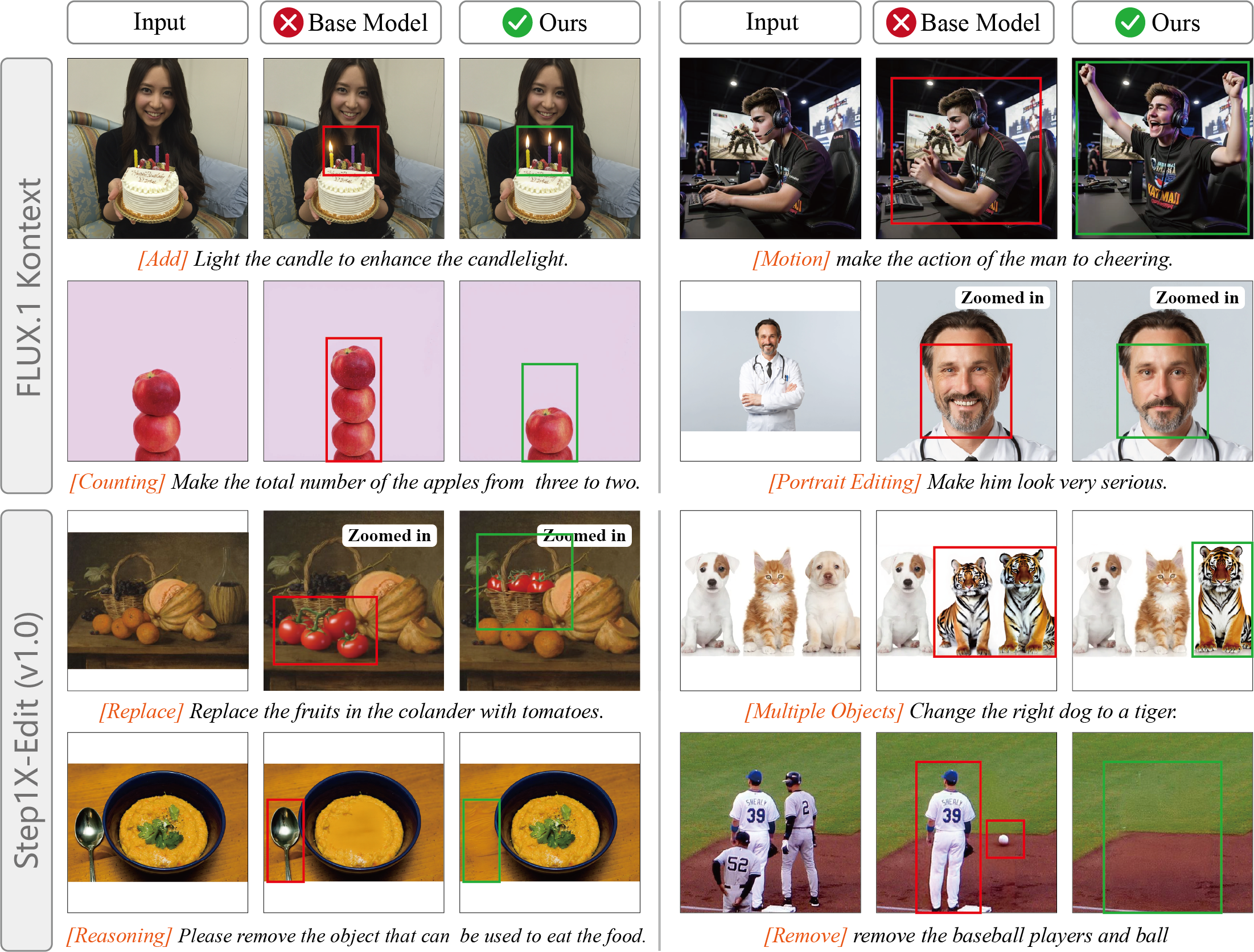}
    \caption{Qualitative results of the base model and VeriLatent across various editing tasks. Task types are shown in orange text. Red boxes highlight incorrect edits produced by the base model, whereas green boxes denote correct edits generated by VeriLatent. These results demonstrate that VeriLatent effectively handles challenging scenarios, including edits requiring precise perception, instruction comprehension, implicit reasoning, and large-motion edits.
    }
    \label{qualitative}
    \vspace{-0.2cm}
\end{figure*}

\subsection{Quantitative Evaluations}
We compare the proposed VeriLatent with recent inference-time scaling methods, including Best-of-N (BoN) \cite{ma2025inference}, PRM \cite{zhang2025let}, PARM \cite{zhang2025let}, TTS-EF \cite{zhang2026enabling} and ADE-CoT \cite{qu2026scale}. Following the experimental setting of ADE-CoT \cite{qu2026scale}, we adopt Qwen-VL-MAX with VIE-Score \cite{ku2024viescore} as the general MLLM verifier in experiments. To ensure a fair comparison, our method uses this general verifier for final selection among the retained candidates. As demonstrated in Table~\ref{tab:main_comparison}, under the upper budget of $N=32$, our method achieves improvements on most metrics across two base models and three benchmarks. For GEdit-Bench-EN \cite{liu2025step1x}, VeriLatent improves both semantic consistency G\_SC and overall performance G\_O. The improvement is particularly pronounced on FLUX.1 Kontext. VeriLatent outperforms the current state-of-the-art method ADE-CoT by $\bf 0.412$ and $\bf 0.387$ on G\_SC and G\_O, respectively. For AnyEdit \cite{yu2025anyedit} , the proposed method achieves comparable performance to ADE-CoT while requiring lower computational cost. As for Reason-Edit \cite{huang2024smartedit}, VeriLatent achieves state-of-the-art performance on all metrics, demonstrating its ability to understand and handle complex editing instructions. Searching in the instruction space helps the model better capture the editing intent and generate results that are more consistent with complex instructions. Furthermore, the proposed VeriLatent significantly reduces computational cost across all three benchmarks. Taking GEdit-Bench \cite{liu2025step1x} as an example, compared with BoN, VeriLatent reduces the NFE by $\bf 81.9\%$ for FLUX.1 Kontext and $\bf 82.6\%$ for Step1X-Edit. These results validate the effectiveness of the proposed early-step latent verifier and adaptive search strategy. By evaluating candidate noises with only 6 steps, the early-step latent verifier substantially lowers the verification cost. The adaptive search strategy further improves efficiency through difficulty-aware budget allocation.

\subsection{Qualitative Results}
To assess VeriLatent qualitatively, we present generated results across different editing tasks in Fig.~\ref{qualitative}. By comparing with the base models, the proposed method can effectively evaluate candidate initial noises, thereby improving the performance. Specifically, the examples of {\it add}, {\it replace}, and {\it portrait editing} demonstrate improved perceptual ability, enabling more accurate identification of the objects and regions that require editing. The examples of {\it counting}, {\it multiple objects}, {\it remove}, and {\it reasoning} further demonstrate stronger understanding of complex instructions, especially those involving multiple objects and implicit semantics. In addition, the {\it motion} example shows that the proposed method can better support larger structural and pose changes, rather than being limited to minor local modifications. These results indicate that VeriLatent is effective across different editing challenges, including target perception, instruction understanding, implicit reasoning, and large motion editing.

\subsection{Ablation Studies}
We conduct ablation studies on GEdit-Bench-EN with FLUX.1 Kontext \cite{batifol2025flux} as the base model to validate the effectiveness of each key component in VeriLatent. The results are reported in Table~\ref{tab:ablation}. Specifically, we first evaluate the variant without final selection. In this setting, we still retain $K$ candidates, rather than a single candidate, for subsequent denoising stage, but remove the final assessment and selection over the decoded results. This ablated variant shows degraded performance, with G\_SC decreasing from $7.637$ to $7.334$ and G\_O decreasing from $7.082$ to $6.860$. This indicates that although the early-step latent verifier can effectively prune low-quality candidates at an early stage, the final image-level selection further facilitates the identification of the best result among the retained candidates. In addition, we evaluate the variant without instruction reformulation. Under the same maximum search budget, this variant also yields lower G\_SC and G\_O scores than the full VeriLatent. This demonstrates the effectiveness of exploring the joint space of initial noises and editing instructions. Reformulating the original instruction into clearer and more executable editing commands helps the model better interpret and execute complex or ambiguous editing instructions. Overall, these results show the contribution of both final selection and instruction reformulation to the proposed method.

\begin{table}[t]
    \centering
    \caption{Ablation study of VeriLatent. We conduct experiments on GEdit-Bench-EN using FLUX.1 Kontext as the base model.}
    \label{tab:ablation}
    \fontsize{5.0pt}{6.0pt}\selectfont
    \setlength{\tabcolsep}{2.8pt}
    \renewcommand{\arraystretch}{1.10}
    \resizebox{\columnwidth}{!}{
    \begin{tabular}{l | c | cccc}
    \toprule
    \multicolumn{1}{c|}{\multirow{2}{*}{\textbf{Model}}}
    & \multirow{2}{*}{$N$}
    & \multicolumn{4}{c}{\textbf{GEdit-Bench-EN}~\cite{liu2025step1x}} \\
    \cmidrule(lr){3-6}
    &
    & G\_SC
    & G\_PQ
    & G\_O
    & NFE$\downarrow$ \\
    \midrule
    
    \rowcolor{baseindigo}
    \textbf{FLUX.1 Kontext~\cite{batifol2025flux}}
    & -- & 6.546 & 7.615 & 6.104 & -- \\
    
    w/o final selection
    & 32 & 7.334 & 7.605 & 6.860 & \textbf{161} \\
    
    w/o instruction reformulation
    & 32 & 7.318 & \textbf{7.693} & 6.817 & 167 \\
    
    VeriLatent full
    & 32 & \textbf{7.637} & 7.663 & \textbf{7.082} & 162 \\
    
    \bottomrule
    \end{tabular}
    }
\end{table}

\begin{table}[t]
    \centering
    \caption{Impact of the maximum budget $N$. Experiments are conducted on GEdit-Bench-EN using FLUX.1 Kontext as the base model.}
    \label{tab:impact_budget}
    \fontsize{5.0pt}{6.0pt}\selectfont
    \setlength{\tabcolsep}{3.8pt}
    \renewcommand{\arraystretch}{1.08}
    \resizebox{\columnwidth}{!}{
    \begin{tabular}{l | c | cccc}
    \toprule
    \multicolumn{1}{c|}{\multirow{2}{*}{\textbf{Model}}}
    & \multirow{2}{*}{$N$}
    & \multicolumn{4}{c}{\textbf{GEdit-Bench-EN}~\cite{liu2025step1x}} \\
    \cmidrule(lr){3-6}
    &
    & G\_SC
    & G\_PQ
    & G\_O
    & NFE$\downarrow$ \\
    \midrule

    \rowcolor{baseindigo}
    \textbf{FLUX.1 Kontext~\cite{batifol2025flux}}
    & -- & 6.546 & 7.615 & 6.104 & -- \\

    \multirow{3}{*}{w/ VeriLatent (Ours)}
    & 8  & 7.506 & \textbf{7.721} & 6.997 & \textbf{136} \\
    & 16 & 7.554 & 7.716 & 7.044 & 148 \\
    & 32 & \textbf{7.637} & 7.663 & \textbf{7.082} & 162 \\

    \bottomrule
    \end{tabular}
    }
    \vspace{-0.5cm}
\end{table}

\subsection{Impact of Maximum Search Budget}
\begin{figure}[t]
    \centering
    \includegraphics[width=7.2cm]{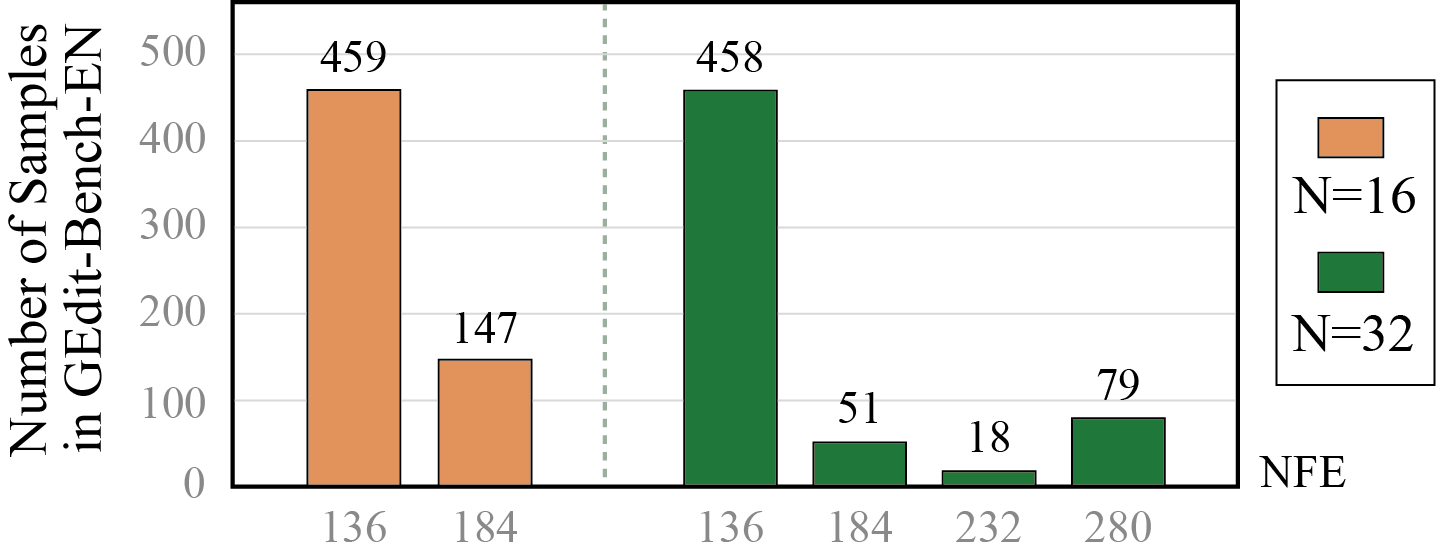}
    \caption{Distribution of GEdit-Bench-EN samples with respect to NFE under different maximum budgets. The majority of samples require only one search round, corresponding to an NFE of 136. The remaining difficult samples require more computation. }
    \label{budget_diagram}
    \vspace{-0.5cm}
\end{figure}

We further analyze the impact of the maximum search budget $N$ on GEdit-Bench-EN \cite{liu2025step1x} with FLUX.1 Kontext \cite{batifol2025flux} as the base model. As shown in Table~\ref{tab:impact_budget}, increasing $N$ consistently improves the semantic consistency G\_SC and overall performance G\_O. Specifically, when $N$ increases from 8 to 32, G\_SC improves from $7.506$ to $7.637$, and G\_O improves from $6.997$ to $7.082$. Under our experimental setting, a larger search budget provides more candidate initial noises, thereby increasing the chance of finding better editing trajectories. More importantly, with the proposed adaptive search strategy, NFE does not grow linearly with $N$. Even when the upper search budget is substantially increased, the average NFE remains at a relatively low level. We further report the distribution of samples in GEdit-Bench-EN with respect to NFE in Fig.~\ref{budget_diagram}. GEdit-Bench-EN contains 606 samples in total. For most samples, promising initial noises can be found after only one search round, corresponding to an NFE of 136. VeriLatent does not allocate excessive computation for these samples. Therefore, increasing the upper search budget does not substantially increase the average NFE, as the adaptive search strategy stops early once enough promising candidates have been found. These results suggest that the proposed adaptive search strategy can effectively allocate computation budget according to the difficulty of the editing task.

\begin{table*}[t]
    \centering
    \scriptsize
    \caption{Comparison between base models and VeriLatent across different editing tasks on GEdit-Bench-EN. We report category-wise G\_SC scores for FLUX.1 Kontext and Step1X-Edit with and without VeriLatent. Values in parentheses denote relative improvements over the corresponding base model. VeriLatent improves G\_SC across all 11 editing tasks for both base models.}
    \label{tab:different tasks}
    \setlength{\tabcolsep}{2.2pt}
    \renewcommand{\arraystretch}{1.22}
    \begin{tabular*}{\textwidth}{@{\extracolsep{\fill}} c|ccccccccccc @{}}
    \toprule
    \multirow{2}{*}{\textbf{Model}} 
    & \textbf{Background} 
    & \textbf{Color} 
    & \textbf{Material} 
    & \textbf{Motion} 
    & \textbf{Portrait} 
    & \textbf{Style} 
    & \textbf{Subject} 
    & \textbf{Subject} 
    & \textbf{Subject} 
    & \textbf{Text} 
    & \textbf{Tone} \\
    & \textbf{Change} 
    & \textbf{Alter} 
    & \textbf{Alter} 
    & \textbf{Change} 
    & \textbf{Editing} 
    & \textbf{Transfer} 
    & \textbf{Addition} 
    & \textbf{Removal} 
    & \textbf{Replace} 
    & \textbf{Change} 
    & \textbf{Transfer} \\
    \midrule
    FLUX.1 Kontext
    & 7.725 & 9.000 & 6.150 & 3.025 & 3.129 & 6.650 & 7.833 & 7.491 & 7.067 & 6.606 & 7.325 \\
    \textbf{w/ VeriLatent}
    & \makecell[c]{\textbf{8.800}\\[-1pt]\textcolor{gainred}{(+13.9\%)}}
    & \makecell[c]{\textbf{9.050}\\[-1pt]\textcolor{gainred}{(+0.6\%)}}
    & \makecell[c]{\textbf{7.525}\\[-1pt]\textcolor{gainred}{(+22.4\%)}}
    & \makecell[c]{\textbf{4.550}\\[-1pt]\textcolor{gainred}{(+50.4\%)}}
    & \makecell[c]{\textbf{5.243}\\[-1pt]\textcolor{gainred}{(+67.6\%)}}
    & \makecell[c]{\textbf{7.567}\\[-1pt]\textcolor{gainred}{(+13.8\%)}}
    & \makecell[c]{\textbf{9.283}\\[-1pt]\textcolor{gainred}{(+18.5\%)}}
    & \makecell[c]{\textbf{8.509}\\[-1pt]\textcolor{gainred}{(+13.6\%)}}
    & \makecell[c]{\textbf{8.300}\\[-1pt]\textcolor{gainred}{(+17.4\%)}}
    & \makecell[c]{\textbf{7.131}\\[-1pt]\textcolor{gainred}{(+7.9\%)}}
    & \makecell[c]{\textbf{8.050}\\[-1pt]\textcolor{gainred}{(+9.9\%)}} \\
    \midrule
    Step1X-Edit
    & 7.775 & 7.225 & 8.000 & 3.600 & 5.586 & 8.033 & 7.133 & 6.228 & 8.017 & 8.010 & 6.750 \\
    \textbf{w/ VeriLatent}
    & \makecell[c]{\textbf{8.275}\\[-1pt]\textcolor{gainred}{(+6.4\%)}}
    & \makecell[c]{\textbf{8.725}\\[-1pt]\textcolor{gainred}{(+20.8\%)}}
    & \makecell[c]{\textbf{8.300}\\[-1pt]\textcolor{gainred}{(+3.8\%)}}
    & \makecell[c]{\textbf{4.500}\\[-1pt]\textcolor{gainred}{(+25.0\%)}}
    & \makecell[c]{\textbf{6.414}\\[-1pt]\textcolor{gainred}{(+14.8\%)}}
    & \makecell[c]{\textbf{8.483}\\[-1pt]\textcolor{gainred}{(+5.6\%)}}
    & \makecell[c]{\textbf{8.950}\\[-1pt]\textcolor{gainred}{(+25.5\%)}}
    & \makecell[c]{\textbf{8.053}\\[-1pt]\textcolor{gainred}{(+29.3\%)}}
    & \makecell[c]{\textbf{8.600}\\[-1pt]\textcolor{gainred}{(+7.3\%)}}
    & \makecell[c]{\textbf{8.576}\\[-1pt]\textcolor{gainred}{(+7.1\%)}}
    & \makecell[c]{\textbf{8.150}\\[-1pt]\textcolor{gainred}{(+20.7\%)}} \\
    \bottomrule
    \end{tabular*}
\end{table*}

\begin{figure}[t]
    \centering
    \includegraphics[width=8.8cm]{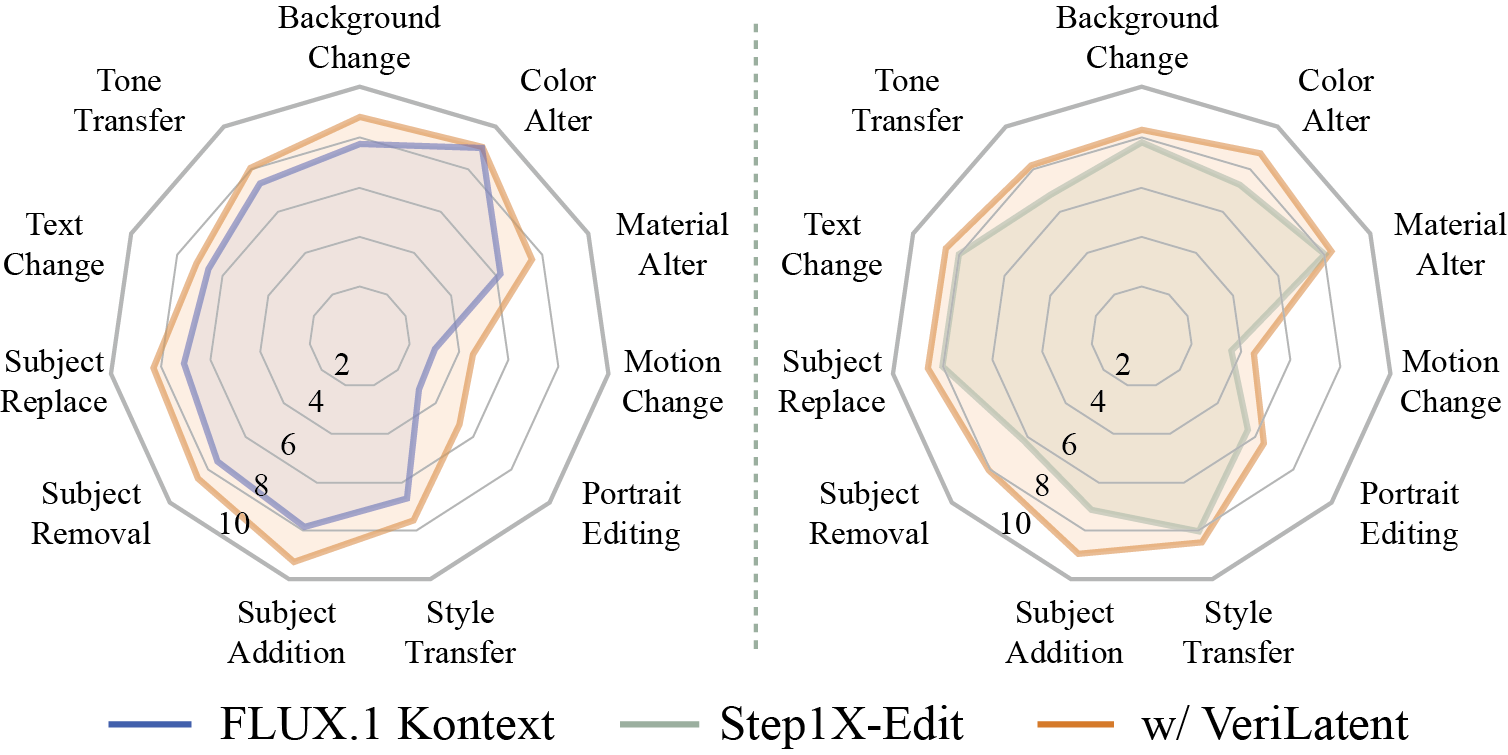}
    \caption{Comparison between base model and VeriLatent in terms of G\_SC across 11 editing tasks on GEdit-Bench-EN. A larger distance from the center indicates a higher score. VeriLatent consistently expands the radar area for both base models, showing consistent improvements across different task types.}
    \label{radar}
\end{figure}

\subsection{Comparison with base model across different tasks}
To analyze the improvements brought by the proposed method across different types of editing tasks, we further compare our method and the corresponding baselines on GEdit-Bench-EN, which contains 11 editing categories. G\_{SC} is utilized to directly evaluate instruction-following ability. As shown in Table~\ref{tab:different tasks} and Fig.~\ref{radar}, our method achieves consistent improvements over two different base models across all the editing categories. Specifically, with FLUX.1 Kontext as the base model, our method yields over 10\% gains in 8 editing categories. For the {\it motion change} and {\it portrait editing} tasks, which are challenging for the base model, our method achieves significant improvements. G\_SC increases by $50.4\%$ and $67.6\%$, respectively. Such improvements show that, via the proposed search strategy, our method can obtain more instruction-consistent results for complex tasks. Additionally, notable relative gains are also observed in {\it material alter}, {\it subject addition} and {\it subject replace}, reaching $22.4\%$, $18.5\%$, and $17.4\%$, respectively. These tasks require the model to accurately identify the target editing region, suggesting that the proposed early-step latent verifier can effectively select initial noises that induce edits in the correct regions. As for Step1X-Edit(v1.0) , our method also achieves consistent improvements. Similarly, for region-specific editing tasks, such as {\it subject addition} and {\it subject removal}, G\_{SC} significantly improves by $25.5\%$ and $29.3\%$. For the {\it motion change} task, G\_{SC} increases from 3.600 to 4.500, corresponding to a $25.0\%$ gain. Overall, the category-wise results show that our method is not only effective for a specific editing type, but also improves instruction-following ability across diverse editing categories.

\section{Conclusion}
In this work, we present VeriLatent, a plug-and-play inference-time scaling framework for instruction-based image editing that leverages early-step latent verification. The proposed verifier assesses initial noises in latent space, efficiently identifying promising initial noises without decoding latents into images. Building upon this, the adaptive search strategy explores the joint space of initial noises and editing instructions, allowing the framework to handle complex editing tasks effectively. Extensive experiments on multiple benchmarks and base models demonstrate that VeriLatent consistently improves performance across diverse editing tasks while significantly reducing computational cost.



\bibliographystyle{IEEEtran}
\bibliography{refs}

\newpage

\section{Biography Section}

\begin{IEEEbiography}[{\includegraphics[width=1in,height=1.25in,clip,keepaspectratio]{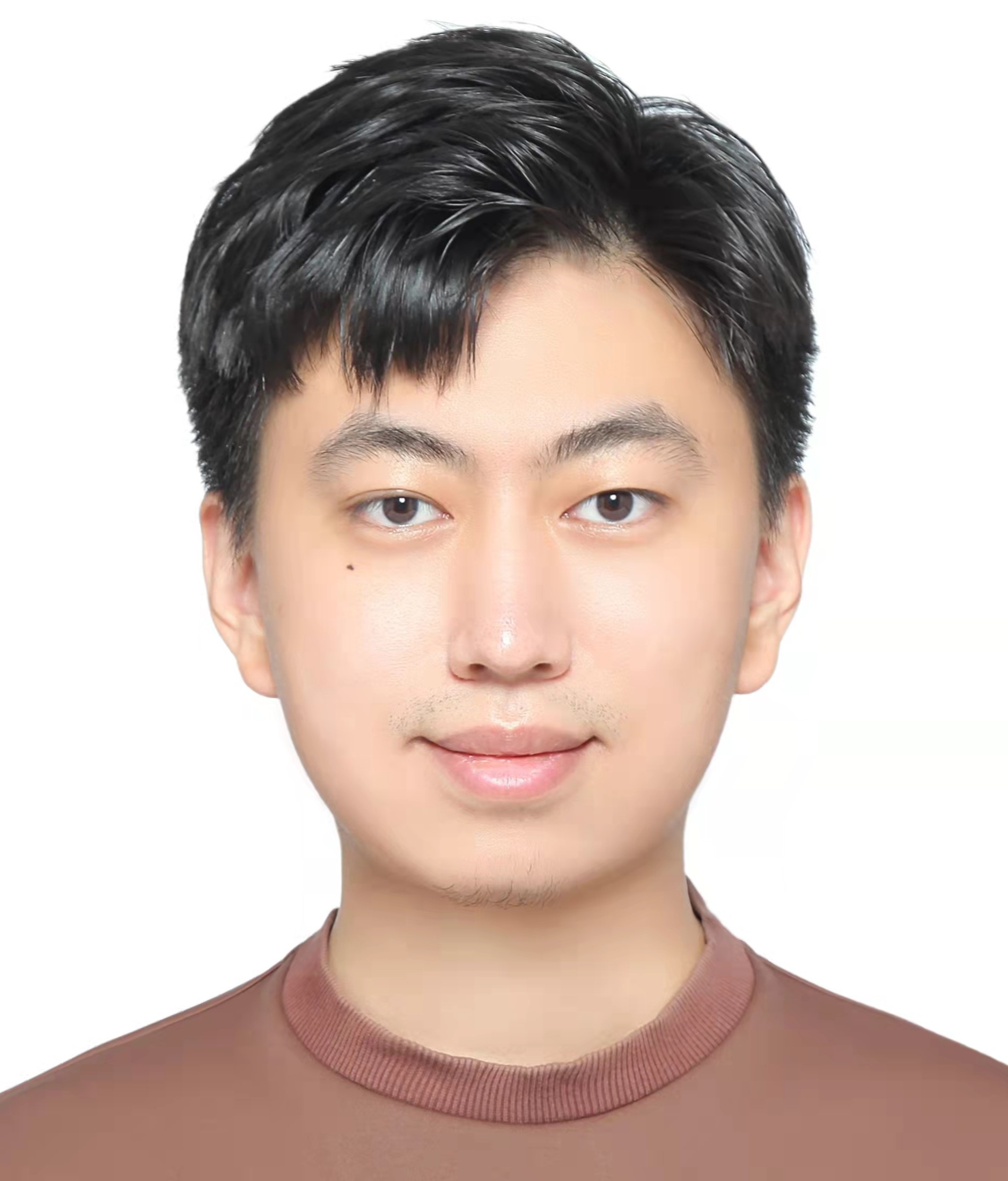}}]{Yue Yu} 
received the Bachelor's degree and M.S. degree from the Chinese University of Hong Kong, Hong Kong, China, in 2019 and Fudan University, Shanghai, China, in 2024. He is currently pursuing his Ph.D. degree in the College of Computer Science and Artificial Intelligence, Fudan University. His research interests include image generation and image editing.
\end{IEEEbiography}

\vspace{-5mm}

\begin{IEEEbiography}[{\includegraphics[width=1in,height=1.25in,clip,keepaspectratio]{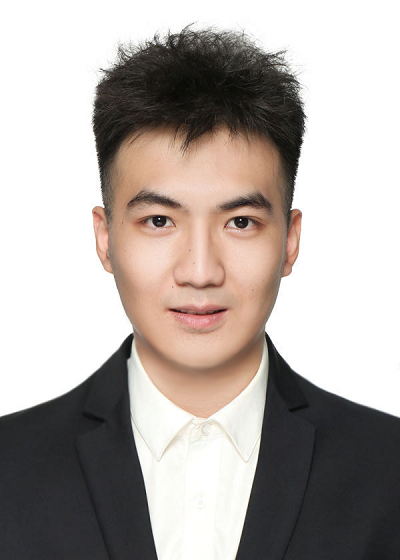}}]{Yang Jiao} received the B.E. degree from the University of Electronic Science and Technology of China, Chengdu, China, in 2021. He is currently pursuing his Ph.D. degree in Computer Science at Fudan University. His research interests include multimedia analysis and 3D vision.
\end{IEEEbiography}

\vspace{-5mm}

\begin{IEEEbiography}[{\includegraphics[width=1in,height=1.25in,clip,keepaspectratio]{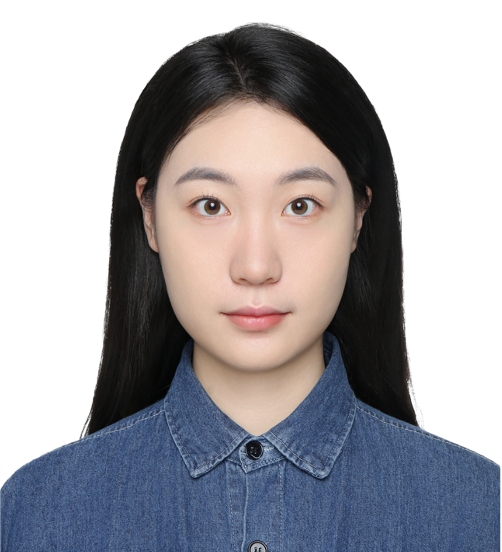}}]{Jiayu  Wang} 
received the B.E. degree from Northwest University, Xi'an, China, in 2022. She is currently pursuing her Ph.D. degree in the College of Computer Science and Artificial Intelligence, Fudan University, Shanghai, China. Her research interests include image generation and image editing.
\end{IEEEbiography}

\vspace{-5mm}

\begin{IEEEbiography}[{\includegraphics[width=1in,height=1.25in,clip,keepaspectratio]{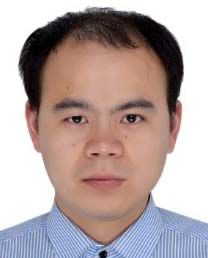}}]{Qi Dai} 
received the Ph.D. degree in computer science from Fudan University, Shanghai, China, in 2017. He is currently a principal researcher at Microsoft Research Asia, Beijing, China. His research interests focus on computer vision and multimedia, with an emphasis on video understanding, analytics, and generation.
\end{IEEEbiography}

\vspace{-5mm}

\begin{IEEEbiography}[{\includegraphics[width=1in,height=1.25in,clip,keepaspectratio]{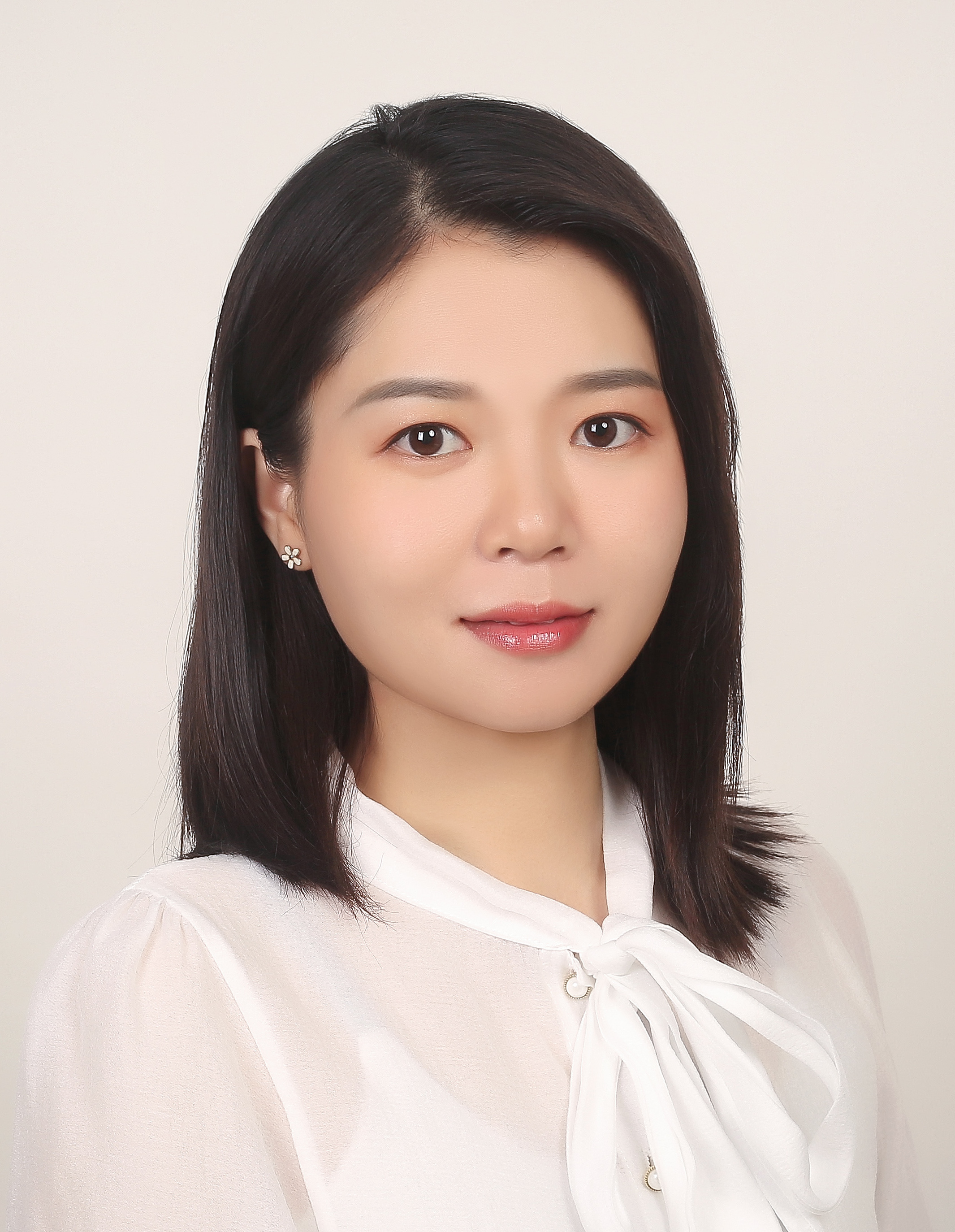}}]{Jingjing Chen} (Senior Member, IEEE) is now an Associate Professor at the Institute of Trustworthy Embodied AI, Fudan
University, Shanghai, China. Before joining Fudan University, she was a postdoc research fellow at the School of
Computing in the National University of Singapore. She received her Ph.D. degree in Computer Science
from the City University of Hong Kong in 2018. 
Her research interests lie in the areas of robust AI, multimedia content analysis, and deep learning. She received multiple prestigious recognitions, including the IEEE Multimedia Rising Star Runner Up Award (2023), the ACM SIGMM Rising Star Award (2024).
\end{IEEEbiography}

\vfill

\end{document}